\documentclass[sigplan,screen]{acmart}

\AtBeginDocument{%
  }

\copyrightyear{2023}
\acmYear{2023}
\setcopyright{rightsretained}
\acmConference[ICDSP 2023]{2023 7th International Conference on Digital
Signal Processing (ICDSP)}{February 17--19, 2023}{Chengdu, China}
\acmBooktitle{2023 7th International Conference on Digital Signal Processing
(ICDSP) (ICDSP 2023), February 17--19, 2023, Chengdu,
China}
\acmDOI{10.1145/3585542.3585546}
\acmISBN{978-1-4503-9862-6/23/02}

\usepackage{amsmath}
\usepackage{graphicx}
\usepackage{float}
\usepackage{hyperref}
\hypersetup{
  colorlinks   = true, 
  urlcolor     = blue, 
  linkcolor    = blue, 
  citecolor   = green 
}
\graphicspath{{images/}} 
\usepackage{algorithm}
\usepackage{algpseudocode}
\usepackage{array, booktabs}
\usepackage[english]{babel}
\usepackage{natbib}
\setcitestyle{square, comma, numbers,sort&compress}
\begin{document}

\title{ Evolution under Length Constraints for CNN Architecture design}

\author{Ousmane YOUME}
\authornote{Principal contributor of this research}
\authornotemark[1]
\affiliation{%
  \institution{Gaston Berger University}
  \streetaddress{Sanar, Saint-louis}
  \city{Saint Louis}
  \country{Senegal}
}
\email{youme.ousmane1@ugb.edu.sn}

\author{Jean Marie Dembele}
\affiliation{%
  \institution{Gaston Berger University}
  \streetaddress{Sanar, Saint-louis}
  \city{Saint-louis}
  \country{Senegal}}
\email{jean-marie.dembele@ugb.edu.sn}

\author{Eugene C. Ezin}
\affiliation{%
  \institution{Abomey-Calavi University}
  \city{Abomey-Calavi}
  \country{Benin}
}

\author{Christophe Cambier}
\affiliation{%
 \institution{Sorbonne University}
 \city{Paris}
 \country{France}}

\renewcommand{\shortauthors}{Youme et al.}

\begin{abstract}
In recent years, the CNN architectures designed by evolution algorithms have proven to be competitive with handcrafted architectures designed by experts. However, these algorithms need a lot of computational power, which is beyond the capabilities of most researchers and engineers. To overcome this problem, we propose an evolution architecture under length constraints. It consists of two algorithms: a search length strategy to find an optimal space and a search architecture strategy based on genetic algorithm to find the best individual in the optimal space. Our algorithms reduce drastically resource cost and also keep good performance. On the Cifar-10 dataset, our framework presents outstanding performance with an error rate of $5.12\%$ and only $4.6$ GPU a day to converge to the optimal individual -22 GPU a day less than the lowest cost automatic evolutionary algorithm in the peer competition.
\end{abstract}


\maketitle
\section{Introduction}\label{sec1}

Among supervised approaches, convolutional neural networks (CNN) are Deep Learning techniques specialized for image processing that have produced unprecedented result in several fields such as medical image analysis, self-driving cars, face recognition, language translation \cite{liu_survey_2017}. The deep structure of CNN allows the model to learn increasingly abstracted layers of descriptors, resulting in representations that can improve the performance of classifiers. CNN can be viewed as a multi-layer perceptron stack whose purpose is to process large amounts of information. Because of their complexity and high dimension, CNN requires training several learnable parameters despite the weight sharing technique to reduce the number of free parameters. The architecture complexity must match with problems to have a suitable model generalization. Let assume the data are too large and complex and the number of learnable parameters of the architecture is small. In that case, we have many solutions in the space that do not have an excellent knowledge of the data. On the other hand, if the data are too small and not great quality, we see a lousy generalization of the model when we have many learnable parameters. Then it is crucial to find an arrangement between the complexity of the model and data. They are two types of hyperparameters for convolution neural networks: hyperparameters related to the structure of the network and hyperparameters related to controlling the training of model. Most of them have the role of regularization to keep the balance between the complexity of the problem and algorithm to avoid overfitting and underfitting. We can formulate the relation of architecture, model and data as follow:

\begin{align}
M &= A(X^{tr} , \lambda)   \label{eq1}
\\
W &= G(A).
\\
\lambda^* &= arg_\lambda minF(\lambda ,A,X^{tr},X^{te},\tau)
\end{align}

Where M is the model obtained from the chosen architecture A computed with $\mathrm{}{X^{tr}}$ the training data. The associated loss function is $\mathbf{\tau}({X}^{te},{M}).$ The model's weights $\mathbf{W}$ are initialized according to the chosen architecture. Finding the optimal model $\mathbf{M^*}$ consists of searching the set of hyperparameters $\mathbf{\lambda^*}$ that minimizes the loss function \cite{claesen_hyperparameter_2015}. Hyperparameter optimization methods are proposed with fixed length (Swam optimization \cite{meissner_optimized_2006}, racing optimization \cite{birattari} etc.). Others methods do not present a good scalability and accuracy (Bayesian optimization \cite{snoek_practical_2012-1}). For the learnable parameters, we have the Gradient Descent based methods which minimize the loss function by optimizing the weights i.e Stochastic Gradient Descent (SGD) \cite{amari_backpropagation_1993},  Adam Gradient \cite{kingma} etc. SGD is scalable and presents a good optimization of the weights. 

However, good architecture design remains the main challenge.  The first successful one is Lenet5 \cite{lecunnet}. Since then, a lot of CNN architectures have been developed Alexnet \cite{krizhevsky_imagenet_2017},  VGG16 \cite{Simonyan}, inception \cite{szegedy_going_2015}, Resnet \cite{he_deep_2015}. These architectures have parallelism, grouped, concatenation, depth and width features.

Recently, algorithms for automating the search of best architecture are proposed in the state of the art with three main methods: a search space strategy, a search for the best architecture and an optimization method to gain computational resources. The first group of algorithm proposed uses Reinforcement learning to design the best architecture MetaQNN \cite{baker_designing_2017}, Neural Architecture Search NAS with reinforcement learning \cite{zoph_neural_2017}. The second group of authors based on first like NAS \cite{zoph_learning_2018} that proposed a search space strategy where each layer of the architecture is predicted based on the performance of its previous. We have also DART \cite{liu_darts_2019} a NAS based gradient descent instead of controller and NAT Neural Architecture Transfer \cite{lu_neural_2021}. These efficient methods require a lot of computational resources running around 2500GPU a Day. It's essential for this algorithm to finish all computation for finding the best architecture as each layer is predicted depending on previous layers. On the other hand, we have genetic algorithm methods  less constraining and easier to manipulable according to your objective.  (Lui et al. \cite{liu_hierarchical_2018}, Real et al. \cite{real_large-scale_2017-1}, Sun et al. \cite{sun_evolving_2019}).
However, these methods also require some computational resources. Therefore, we have to find the best individual by optimizing the search to be computationally efficient.

Our method is based on two algorithms: a search length strategy which consists of searching the optimal space where the best architecture is expected to be found given the length and dataset in order to drastically reduce the search space and a genetic algorithm for finding the best individual in this space. With Cifar 10 in our benchmark, we obtain +1,49 error rate compared of our best peer competitor and 4.6 GPU a day, less costly than peer competitor.

\section{Background and Related Work}\label{sec2}
\subsection{Convolutional Neural Network }\label{subsec12}
CNN is a Deep Learning method that has shown satisfactory performance in processing two-dimensional data with grid-like topology, such as images and videos. CNN consists of a sequence of layers, where every layer transforms the activation or outputs of the previous layer through another differentiable function. It's composed of multiple blocks. Layer types include convolutional layers, pooling layers, and fully connected layers. Convolutional layers transform input images into multiple feature maps by transformation with feature kernels. Pooling layer perform dimension reduction of the feature map. The last layers is flatten into vectors by fully connected layers.

\subsubsection{Convolutional layers }

The convolutional layer computes the convolutional operation of the input images using kernel filters to extract fundamental features. Through a Jacobian transformation of the matrices, the filters allow finding invariant. Filters are convolved with input resulting to a feature map or with the feature map of prior layers. This scalar product is done from left to right from top to bottom with k-pixel stripes which in turn undergo the same process. All features map resulting from each filter of same layer level are superimposed  along the depth dimension.

The application of the filter in part of the input allows to extract as much as possible features in a local region in order to learn features representation that best fit local context. Furthermore the filter parameters are shared for all local positions. Filters transfer the resultants of the weights  after computing. Weight sharing reduces the number of variables affecting learning efficiency and good generalization. Here are some convolution layer types: simple convolution\cite{lecunnet} i.e Conv($3\times3$,Numbers of filters), $1\times1$ convolutions \cite{lin} , flattened convolution \cite{jin} i.e $C\times1\times1$, Spatial and Cross-Channel convolutions \cite{szegedy_going_2015}, Depth-wise Separable Convolutions \cite{chollet}, Grouped Convolutions \cite{xie_aggregated_2017}, Shuffled Grouped  Convolutions\cite{zhang_shufflenet_2017}. Because the convolution is structured so that the reduction of dimensions is limited with simple convolution, the complexity of learning with thousands of free parameters is always present, hence CNN present pooling layer. 

\subsubsection{Pooling (max pooling, average pooling) }
By pooling layers, the number of trainable parameters for subsequent layers is reduced. This reduces a part $k\times k$ after a transformation of the previous matrix $M\times N$ by taking the maximum or averaging at this location. Pooling is used to obtain invariant to the translation of the input over a local neighborhood \cite{boureau_theoretical_nodate-1}\cite{boureau_ask_2011}. The most significant goal of pooling is to transform feature maps into new feature maps with only significant information, discarding details in order to avoid overfitting. The network is also robustness against noise. At the end a Fully connected network combines the last matrices into a vector that will be taken as input in one or more dense neural networks. Apart from weight sharing, translation and invariant, CNN architecture also has equivariant property: meaning that whenever the input changes, the output changes in the same way.

 \subsection{Related work }\label{subsec22}
For decades, evolutionary algorithms have been used to build a dense neural network: neuroevolution. This field began in 1986 with  \cite{miller_designing_1989} then \cite{stanley_evolving_2002}, \cite{bayer_evolving_2009},  \cite{stanley_hypercube-based_2009}. The application of reinforcement learning in this field has led to advancements in the field of robotics and video games \cite{such}. In order to make neurons evolve, these methods rely on the density and connections between neurons whereas with the CNN the layers are linked indirectly via interaction filters.

The evolutionary algorithms have been recently introduced to the CNN with genetic algorithm GA. \cite{real_large-scale_2017} show that it is possible to evolve models with the accuracy of those recently proposed in S.O.T.A. using a basic evolution technique on a large scale. Their algorithms LEIC explore a vast space of search for the best individual starting from zero layers of input-identity-output. Individuals are encoded with layers of variable parameters that are incorporated as they evolve using the mutation method to converge to the best individual. The authors have also studied in this paper the impact of the population size and the number of training steps per individual in order to respectively avoid the trapped population and accelerate the evolution with weight inheritance and not retrain at the end the best model. The crossover that allows local search in GA is not used in the principal part of LEIC but on additional experiments to inherit the right choices of mutations, inheritance weights of dual parents and fusion of two parents side by side. Hence to converge to the best individual this algorithm needs a lot of resources with a population of size $\mathrm{}{10^{3}}$ generally $\mathrm{}{10^{2}}$ in GA trained on 25600 steps with $250$ latest generation machines. This algorithm consumes $2500$ GPU a day under the whole evolution. 

\cite{sun_automatically_2020} propose an AE-CNN framework : a genetic algorithm based on ResNet \cite{he_deep_2015} and Densenet\cite{huang_densely_2017} block. A gene encoding strategy is used to initialize a population and the evolution of individuals begins until the satisfaction criteria or the number of generations are reached. Individuals are encoded by choosing from Resnet Block RB or Densnet Block DB or pooling units of length $k$ randomly selected in order to generate model CNN with different architectures . During evolution, each individual's fitness is estimated based on the accuracy of classification in the validation dataset. Then mutation and crossover methods are proposed for global and local search for the best individual. This algorithm requires 36 GPU days to converge to the best individual. Because of its flexibility and adaptability, we continue to use same mutation and crossover methods for our study. Other proposals have been proposed in the literature semi-automatic search \cite{xie2017genetic} \cite{liu_hierarchical_2018} and whole automatic  evolution \cite{zoph_learning_2018} \cite{baker_designing_2017}.
\section{Algorithms}\label{sec3}
Our method is based on two algorithms. First, a proposal of a search length strategy that returns the optimal space $S$. Secondly, from the beginning of the optimal space $S_{min}$, a Search Architecture Strategy evolves and generates a population of $N_i$ individuals with mutation and crossover methods. We expect to find the best architecture given the length $L$ and dataset $D$ in the optimal space $S$. An individual is composed of blocks of genes. One block has various layers: a convolutional layer, a pooling layer and fully connected layers in a predetermined disposition. The first blocks of size max $3$ are composed of Conv-layers followed by pool-layers i.e Block 1:(Conv-Conv-Pool). The last one is composed of dense layers. We add batch-normalization \cite{ioffe2015batch} and dropout \cite{hinton2012improving} disposition in order to avoid overfitting of the model.

 Our hypothesis consists of:
\begin{enumerate}
\item For the evolutionary research, best individual can be find knowing the model $M$ is a function of architecture $A$ and data $D$ (equation \ref{eq1}). Then the solution fixes the most this equation is on scale order(see sub-section {\ref{subsec32}}).
\begin{quote}
Let the distance between two candidates in two spaces according to the length is $k$ layers. The best model is in $k$-best-space such that
$M(k_{bestspace}) - M(k_{space}) \ge \alpha$ \label{eq4}
for all $k$ in range of spaces list and $k$ different of $k$-best-space.
\end{quote}
\item Therefore, we find the optimal space, we generate $N_i$ individuals with different initialization, and apply mutation method for global search and crossover method for local search throughout the evolution \cite{sampson1976adaptation} to converge to the best individual.
\end{enumerate}
\subsection{Search Architecture Strategy }\label{subsec3}

\begin{algorithm}
	\caption{Search Architectures Strategies}\label{algo1}
	\begin{algorithmic}[1]
	    \State L = SearchLengthStrategy()
	    \State $\mathbf{P}_{0} \longleftarrow$  Initialize population on the given length L
	    \State $t \Leftarrow 0$
		\While{Criteria not satisfied and $t \leq N_G $}
		\State Evaluate I in $\mathbf{P}_{t}$
		\For{Individual $I_1$ $I_2$ Selected on $\mathbf{P}_{t}$}
		\If{Probability $P_1$ greater than $P_c$}
		\State  Crossover $I_1 I_2$ 
		\EndIf
		\If{Probability $P_2$ greater than  $P_m$}
		\State Mutation $I_1, I_2$
		\EndIf
		\State Add evaluated individuals on $G_t $
		\EndFor
		\State $\mathbf{P}_{t+1} \longleftarrow $ Select population from $\mathbf{P}_{t} \cup \mathbf{G}_{t}$
		\State $t \longleftarrow t+1$
		\EndWhile
		\State Return the best individual selected
	\end{algorithmic} 
\end{algorithm} 
This algorithm receive the best space returned by the search Length Strategy algorithm developed below \ref{subsec32}. Then, it generates the first generation by having an optimal space of length L. The search architecture strategies algorithm includes all algorithms developed to find the best candidates. Here we designed two algorithms of initialization considering length L. For the first initialization algorithm, the Standard-Generation algorithm generates individual in a range of one to lengths block. These blocks are composed of one Convolution layer followed by one or two pooling layers and at the end two fully connected layers. Our second Random-Generation generates, in the same range, random disposition in blocks beginning conventionally by convolutional layers in CNN. In order to have the best initialization parameters, we have grown from the bottom of the space to the top. The size of individuals varies by one or two layers in initialization in order to avoid the same size for all individuals at the beginning. Our algorithm for determining the optimal space allows us to eliminate several spaces at once. Henceforth, we can decrease the population size $N_i$ and the number of generations $N_g$ \ref{subsec42}. 

When evaluating individuals, accuracy represents their fitness. During the operation of crossover two chromosomes exchange parts of their layer's parameters to give new chromosomes. These interchanges can be single or multiple. In our case, there are several crossing points because besides length, others parameters like weights, stride and kernel are more important so being able to cross from $I_1$ to $I_2$ at multiple points can permits to interchange several initial parameters between individual. The operation of mutation is used to avoid a premature convergence of the algorithm. When searching for an maximum, the mutation avoids convergence to a local maximum. We have two mutation operands. One is add layers into blocks that permits to modify the length so the capability of individual to fit data by increasing its complexity. Add operand is casting such that the length of the individual does not exceed the optimal space. Second, update operand change layers to other types of layers in order also to improve individual by increasing or decreasing complexity (i.e. Conv to Pool).

In genetic algorithms, there are various natural methods of selection: Roulette Wheel Selection, Rank Selection, Steady State Selection, Tournament Selection, Elitism Selection, Boltzmann Selection. We choose Tournament Selection. This technique uses proportional selection on pairs of individuals and chooses from these pairs the individual with the best adaptation score. This technique is used to select individuals from Population $P_t$ for generating new population $G_t$ with the method of crossover and mutation. An environmental selection with Elitism selection is set to select and return the population $P_{t+1}$ of the next generation. We repeat the process until the best individual satisfies the fixed criteria or reaches the size of generation.
\begin{figure}
    \centering
    \includegraphics[width=7.5cm, height=6cm]{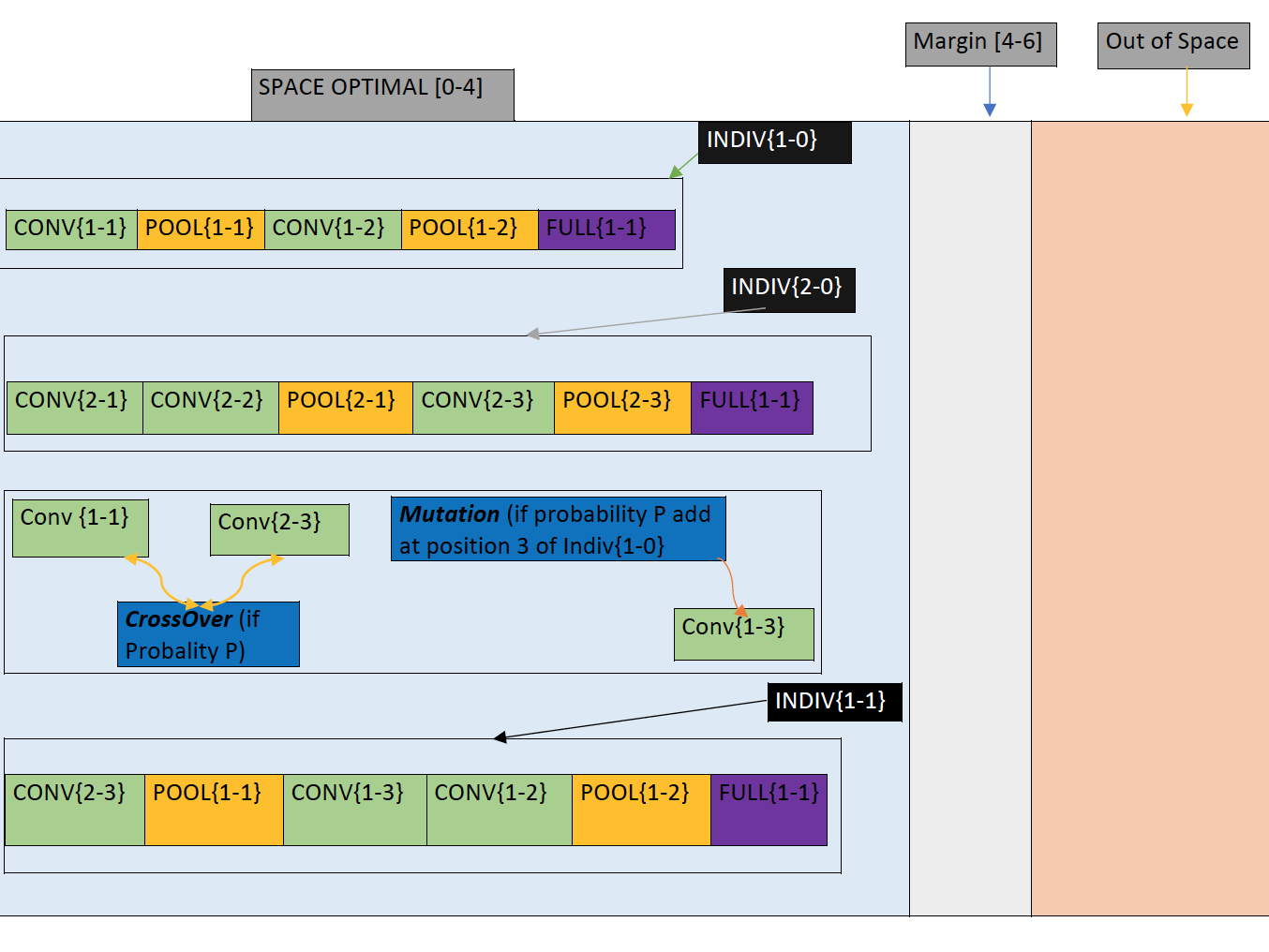}
    \caption{Illustration of application method on two individuals with different lengths. Here we apply on individuals one mutation and crossover methods. The first operation set a crossover method between the first Convolutional layers of individual 1 with the third Convolutional layers of individual 2. Second a mutation method with add operand of a Convolutional layers at position 3 in individual 1. }
    \label{fig1}
\end{figure}
\subsection{Search Length Strategies }\label{subsec32}
 In this section, we developed our main algorithm consisting of searching the optimal space where we expect to find our best individual. First, we gave the number N of maximal layers.
\begin{algorithm}
\caption{Search Length Strategy}\label{algo2}
\begin{algorithmic}[2]
\Require $N $ Number maximal of layers
\Ensure Length L =[min,max] 
\State $S \Leftarrow \emptyset$
\While{$i \leq $N/4}
        \State $\mathbf{n}_{cp} = 2 \times(2 \times i-1)$ 
        \State $Ind \Leftarrow \emptyset$
        \For{$k \leq \mathbf{n}_{cp}/2$ }
        \State Generate $Ind \cup Conv Layers$
        \State Generate $Ind \cup Pooling Layers$
        \EndFor
        \State $ \mathbf{n}_{f} \Leftarrow Generate number between [1, 3]$
        \For{$j \leq \mathbf{n}_{f}$}
        \State $Ind \cup FullConvLayers$
        \State $Batch Normalization$
        \State $Dropout $
        \EndFor
        \State $Ind \cup FullConvLayers$
        \State $E = Evaluate(I, Epoch =5)$
        \State $\mathbf{S}_{1} \Leftarrow{[4(i-1),4i , E]}$
        \State $S = \mathbf{S}_{1} \cup S $
\EndWhile
\State $L \Leftarrow Selection_Space(S)$
\State $Return Length L$
\end{algorithmic}
\end{algorithm}
A distance of $k$ layers exists between the middle of different spaces. The idea's whenever we are in optimal space $S$ and we add or remove $n$ layers that will affect the complexity of architecture. We take the middle of each space as a candidate representing that space. Let $k=4$; we obtain for 24 $N{max}$ layers 6 spaces: {[0-4], [4,8], [8,12], [12,16], [16,20], [20,24]}. Algorithm \ref{algo2} generate a standard individual with random parameters of length $N_{cp} = 2 \times(2 \times i -1)$ corresponding to the length of candidates from space $i$ . According to the standard initialization, the convolutional layer and pooling layer are embedded in a natural arrangement, with the pooling layer following one convolutional layer follow at the end fully connected layers. In order to ensure that random parameters do not mislead the algorithm or give no optimal space when all of them fail, the execution of training candidate model is repeat five ($5$) times. Then, acquired the fitness of representative individuals, the algorithm compared them, select the best spaces, and return at the end the optimal spaces $S$. Let's assume the best individual is at the beginning in $j+1$ space and the fitness of candidate in $j$ space  greater to the fitness of candidate in $j+1$ space then $j$ is selected as best space. To take account of this case, a margin space of $k/2$ layers in algorithm \ref{algo1} is set in order for individuals to be able to touch the next space with the mutation add operand. The parameter $k$ is set depending on the data structure. If the complexity of data is high or the quantity of data is small then we reduce $k$ otherwise k is increased. The algorithm returns no optimal space if all candidates have no good fitness. At times, only dense layers can be sufficient. For this purpose, a zero representative candidate is added to algorithm initialization. According to our result on Cifar-100, several classes can lead the algorithm to go further in-depth of parameters for finding one good or able candidate. In the Next section \ref{sec4}, we report our experimentation in dataset and after in  section \ref{sec5} the results.
\section{Experimentation }\label{sec4}
Here the benchmark data in \ref{subsec41} is run and result compared  in section \ref{sec5} with some powerful CNN algorithms in S.O.T.A. in terms of time cost and precision. In section \ref{subsec42}, all parameters for initialization are presented , such as the size of the population and the maximum length of individual.
\subsection{Benchmark Dataset}\label{subsec41}
For experimentation of our two main algorithms, we have four image classification dataset: Original Mnist, Fashion Mnist, Cifar-10, and Cifar-100. These benchmarks are used to prove efficiency of our algorithms. Resources cost and performance are compared to other algorithms in the state of the art. 

Original Mnist is a large database of handwritten digits created by "re-mixing" the samples from NIST's original datasets\cite{lecunnet}. It contains 60,000 examples of the training set and 10,000 examples of 10 categories between [0-9]. The images have size of $28\times28$ and are in gray scale.
\begin{figure}[H]
    \centering
    \includegraphics[width=7.5cm, height=1.5cm]{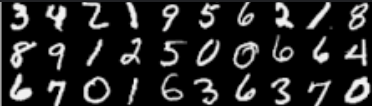}.
    \caption{Extracted images on Mnist Original Dataset.}
    \label{fig2:Mnist}
\end{figure}
The Fashion Mnist shares the same image size and structure of training and testing splits that original Mnist dataset. Each training and test example correspond to one of the following labels: 0 T-shirt/top, 1 Trouser, 2 Pullover, 3 Dress, 4 Coat, 5 Sandal, 6 Shirt, 7 Sneaker, 8 Bag ,9 Ankle boot.
\begin{figure}[H]
    \centering
    \includegraphics[width=7.5cm, height=1.5cm]{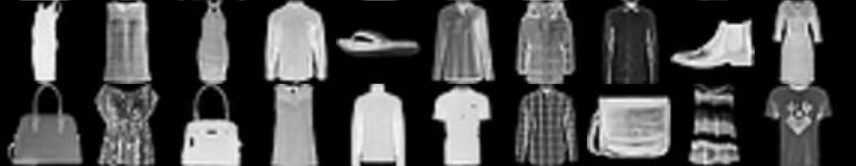}.
    \caption{Image extract on Fashion Mnist Dataset.}
    \label{fig3:Fashion Mnist}
\end{figure}

The Cifar-10 \cite{krizhevsky_learning_2009} is a dataset widely used to evaluate image classification models. Cifar-10 is very useful to validate model of image classification and to compare with peer competitors. It contains 60,000 images of $32\times32\times3$ divided in 10 classes (airplanes, cars, birds, cats, deer, dogs, frogs, horses, ships, and trucks) and  6,000 images for each class.  It is split in two group of dataset 50,000 training images and 10,000 test images. Cifar-100 is a larger format than cifar-10 contains 100 classes divided into 600 images per class. Classes are grouped into 20 super-classes i.e. super-class: fish have 5 classes: aquarium fish, flatfish, ray, shark, and trout.
\begin{figure}[H]
    \centering
    \includegraphics[width=7.5cm, height=1.5cm]{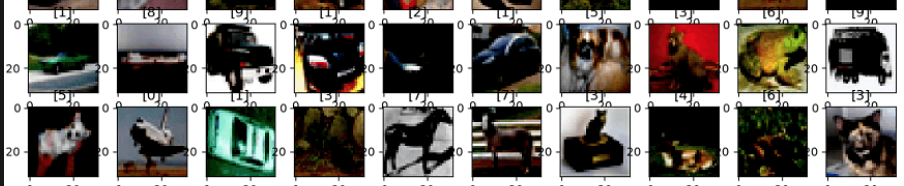}.
    \caption{Image extract in Cifar-100 dataset.}
    \label{fig4}
\end{figure}
\subsection{Parameters Initialization}\label{subsec42}
Since algorithm \ref{algo2} allows to reduce the search space, we reduce at the same time  all the parameters of the genetic algorithm \ref{algo1}.  The population size is set to $25$ individuals and the maximum generations to $10$. The size of the population is significant because a large size explores more space and therefore does not converge to a local optimum. However, reduced space a lot, lead model to start from optimal space. Then it is not mandatory a large population to converge to the best individual. Similarly, the number of generation is reduced. If the algorithm reaches the maximal length in optimal space only crossover and mutation with update operand are applied to individuals. The mutation operand add layers such as the modified length should not exceed the optimal length. The probability of mutation is set to $0.5$ conventionally $0.2$ in genetic algorithm due to the priority of length modification and the probability of crossover to $0.2$ conventionally $0.9$ because it is not desirable to change the initial parameters too much before reaching the maximum size. For our algorithm \ref{algo2}, we set the maximal number of length to $24$ according to the common max length of standard algorithm and the size of $k=4$ (four) layers with a margin of $2$ (two) layers. The alpha difference  for choosing the best between the accuracy of two individuals in two different spaces: $0.05$ for all datasets.

\section{Experimental Results}\label{sec5}
In this section, we show the results obtained from our algorithms  on the benchmark Dataset used. First, we show the efficiency of algorithm \ref{algo2} and secondly the impact of this algorithm on computational resources, accuracy, and loss. After we compare algorithm \ref{algo1} with peer competitors. 

\subsection{Search Length result}\label{51}
\begin{figure}[H]
    \begin{minipage}[c]{.46\linewidth}
        \centering
        \includegraphics[width=4.5cm, height=4.5cm]{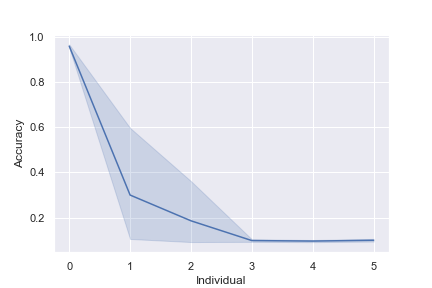}.
    \end{minipage}
    \hfill%
    \begin{minipage}[c]{.46\linewidth}
        \centering
        \includegraphics[width=4.5cm, height=4.5cm]{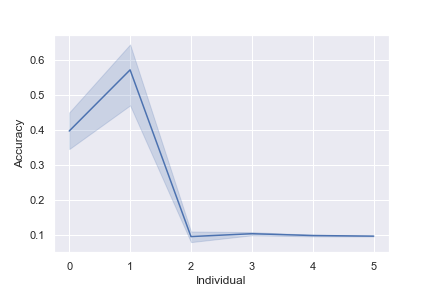}.
    \end{minipage}
    \caption{Variation of fitness for candidates in proposed spaces on: (a) Mnist dataset (b) Cifar-10 dataset.}
    \label{fig5}
\end{figure}

This experiment was run in $20\%$ of dataset in 100 epochs. We note for the dataset of Mnist Original and Cifar-10 a clear limitation of optimal spaces in figure \ref{fig5}

In this figure \ref{fig5} we note an evolution of curves from optimal spaces and a decrease in other spaces for both dataset Mnist original and Cifar with a variation in Mnist. To be sure of the result with random parameters of the model, algorithm is executed $5$ (five) times.

\begin{figure}[H]\label{fig6}
    \begin{minipage}[c]{.46\linewidth}
        \centering
        \includegraphics[width=4.5cm, height=4cm]{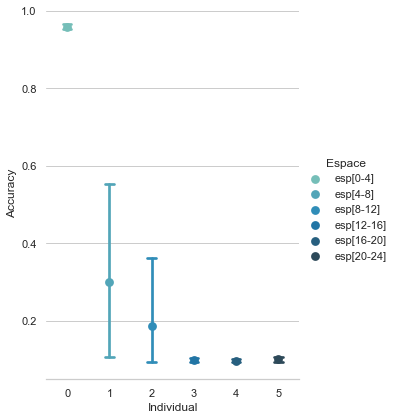}.
    \end{minipage}
    \hfill%
    \begin{minipage}[c]{.46\linewidth}
        \centering
        \includegraphics[width=4.5cm, height=4cm]{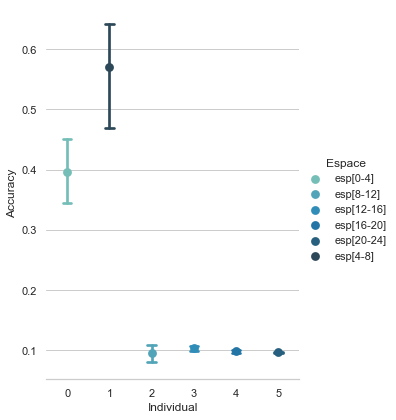}.
    \end{minipage}
    \caption{Result of algorithm Search length Strategy executed five times on: (a) Mnist dataset (b) Cifar-10 dataset.}
\end{figure}

For the Original Mnist Dataset, the fitness of the candidate in the first space [0-4] (the optimal space) is far better than candidates in other spaces. In the optimal space, all candidates have good fitness and the other candidates in the next space exhibit variation. As a result of Cifar-10, the optimal space determined is [4-8] sensibly greater than the candidate spaces of the first space [0-4] and clearly distinguishable from the following spaces, which are near zero due to their out-of-shape resulting from down-sampling. Although the algorithm \ref{algo2} has more difficulty finding an optimal space in Fashion Mnist and Cifar-100. Fashion Mnist's first space has variations based on parameter initialization, and all the other spaces are always worthwhile to explore. Individual generated, with only dense layers, gives higher satisfaction than all other candidates. Because Cifar-100 has many classes and small data for each class, the variation in accuracy between spaces is in order of $0.1$. Nonetheless, the algorithm returns an optimal space with few test losses and better accuracy for each execution.

The mutation and crossover methods in algorithm {\ref{algo1}} lead population in optimal space to converge to best individual.
\begin{figure}[H] \label{fig7}
    \begin{minipage}[c]{.46\linewidth}
        \centering
        \includegraphics[width=5cm, height=4.5cm]{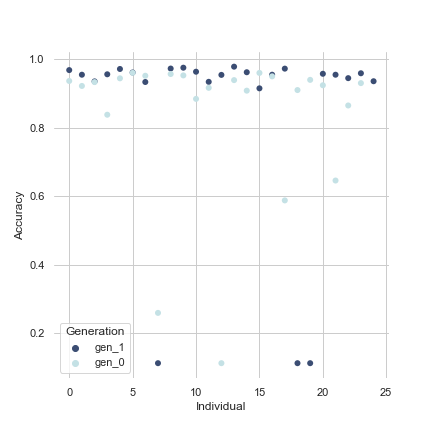}.
    \end{minipage}
    \hfill%
    \begin{minipage}[c]{.46\linewidth}
        \centering
        \includegraphics[width=5cm, height=4.5cm]{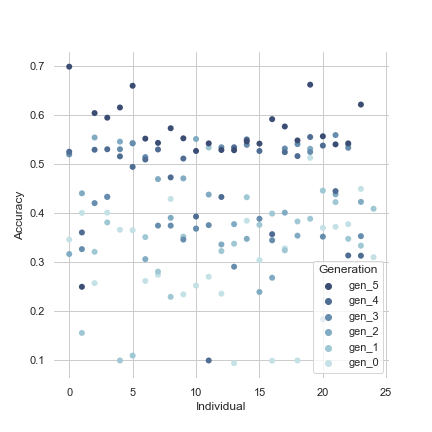}.
    \end{minipage}
    \caption{Impact of Crossover and mutation over generation on (a) Mnist dataset (b) Cifar-10 dataset}
\end{figure}

Figure 7 shows efficiency of these methods before the global selection method with five generations of cifar-10 and one generation in Mnist Original. This confirms that the population is not trapped by evolution and does not converge to a local optimal. Results show a favorable evolution in both datasets Original Mnist and Cifar-10 at the start. 
\subsection{Evolution Result}
In this section, we compare the results obtained in our proposition with peer competitors in the state-of-the-art. In the table \ref{tab2}  are three sections: first a comparison with a hand-crafted algorithm, second a comparison with a semi-automatic algorithm, and finally a comparison with an automatic evolution algorithm regarding accuracy, loss, and GPU cost. GPU/Day is the number of days needed times the number of GPU used to find the best individual. Using the sign $--$ meant that there was no value for this column and using $-$ or $+$ signified the difference between the value and the best noted in the table. The number of parameters for each model is in order of Million M. Cenet models (ours) are trained on the entire datasets of Cifar-10 and Cifar-100 in 400 epochs with 128 batch-sizes. We choose Stochastic Gradient Descent with a learning rate of $10^{-3}$ and momentum of $0.9$.

In the Dataset of Cifar-10 and Cifar-100, we set the number of Generations at $10$ and $25$ individuals in the population. We obtained  $5.12\%$ loss for Cifar-10 computed in $4,6$ GPU day including the cost of algorithm \ref{algo2}: $0.6$ GPU day and $22.16\%$ for Cifar-100 computed in $6,6$ GPU day. Compared to Hand-crafted design, Cenet have $-1.54\%$ than VGG16 \cite{krizhevsky_imagenet_2017} , $-2.81\%$ error rate than Resnet \cite{he_deep_2015} and $-0.12\%$ than Densenet \cite{huang_densely_2017}. In all peer competitors on hand-crafted designs,  Cenet has less error rate except in Wide Resnet \cite{zagoruyko2016wide} with $+0.3\%$ error rate. In the same configuration, a comparison with semi-automatic architecture shows that Cenet has $-1.9\% $less error rate than Genetic CNN but $+1.49\%$ than Hierarchical Evolution \cite{liu_hierarchical_2018} the best peer competitor present here in Cifar-10 dataset but almost 100X less GPU Day. We also gain more computational resources $-5,4$ Gpu Day than EAS \cite{sun_automatically_2020}, $-13$ Gpu Day than Genetic CNN. Cenet present also fewer parameters than all in this section. In category of automatic evolution, our algorithm don't have much variation in error rate $-0.28\%$ than LEIC \cite{real_large-scale_2017} evolved from scraft, $-0.86\%$ than CGP-CNN \cite{suganuma2017genetic}, less than method based reinforcement learning : MetaQNN \cite{baker_designing_2017} $-1.8\%$ and NAS \cite{zoph_learning_2018} $-1.10\%$ also in Cifar-10. However Cenet is worth than EA-CNN \cite{sun_automatically_2020} that we inspired $+0.82\%$ with approximately same numbers of parameters. This is due to Cenet design simpler architecture and not embedded different complex architecture like Densenet block, Resnet block, or transformers encoder. Cenet gets less cost resources consumption than all competitor with $600X$ less GPU a day than LEIC \cite{real_large-scale_2017} and $-22$  GPU a day less than AE-CNN \cite{sun_automatically_2020}. In summary we can say that our algorithm in terms of resource cost consumes less than all the current competitor's evolutionary models and despite the fact that it does not embody different complex architecture maintains a performance close to theirs.


\begin{table*}[h]
\begin{center}
\begin{minipage}{\textwidth}
\caption{Result of evolutionary algorithm selected}\label{tab2}
\begin{tabular*}{\textwidth}{@{\extracolsep{\fill}}lcccccc@{\extracolsep{\fill}}}
\toprule%
& \multicolumn{3}{@{}c@{}}{Cifar10\footnotemark[1]} & \multicolumn{3}{@{}c@{}}{Cifar100\footnotemark[2]} \\\cmidrule{2-4}\cmidrule{5-7}%
Model & GPU/Day & Pm & $\sigma_{loss} $ & GPU/Day & Pm  & $\sigma_{loss}$\\
\midrule
VGG16 \cite{krizhevsky_imagenet_2017}  & - & 20.04 M & $6.66$ & - & - & $28.05$ \\
DenseNet\cite{huang_densely_2017}  & - & 1.0M  & $5.24$  & - & 1.0M & $24.42$\\
ResNet\cite{he_deep_2015} & - & 10.2M & $7.93$  & - & 10.2M & $27.82$\\
Wide ResNet\cite{zagoruyko2016wide} & - & 11M & $4.81$  & - & 11M & $22.89$\\
Maxout\cite{goodfellow_maxout_2013} & - & - & $9.3$  & - & - & $38.6$\\
Network In Network\cite{lin} & - & - & $8.81$  & - & - & $38.6$\\
HighWay Network\cite{srivastava_highway_nodate} & - & - & $7.72$  & - & - & $32.39$\\
\bottomrule
Genetic CNN\cite{xie2017genetic} & 17 & - & $7.1$  & - & - & $29.05$\\
Hierarchical Evolution\cite{liu_hierarchical_2018} & 300 & - & $3.63$  & - & - &$29.05$\\
EAS \cite{cai_efficient_2018}  & 10 &23.4M & $4.23$  & - & - &\\
Block-QNN-S\cite{Sutskever}  &  90 &6.1M & $4.38$  & - & $20.65$&\\
\bottomrule
LEIC\cite{real_large-scale_2017-1} & 2750 & 5.4M  & $5.4$  & 2750 & 40.4M & $23$\\
CGP-CNN\cite{suganuma2017genetic}& 27 & 2.64M & $5.98$\\
NAS\cite{zoph_learning_2018} &22400& 2.5M&$6.01$& - & - &-\\
MetaQNN\cite{baker_designing_2017} &100& - & $6.92$ &-&-&$27.14$ \\
AE-CNN\cite{sun_automatically_2020} & 27 & 2.0M & $4.3$ & 36 & 5.4M &$20.85$ \\
\bottomrule
Ours CENET & $4,6(-5.4)$ & 3.5M & $5.12(+1.49)$  & 5,6(-21) & 5.2M& $-$\\
\end{tabular*}
\footnotetext{Note: Result of competitor has been extracted from original paper.}
\footnotetext[1]{ 1: Result on Cifar 10 Dataset.}
\footnotetext[2]{ 2 :Result on Cifar 100 Dataset.}
\end{minipage}
\end{center}
\end{table*}

\section{Further Discussions}\label{sec6}

From the comparison of a peer competitor, we can affirm now that our algorithms keep a balance between resources cost much less than all competitors and also perform less error than many competitors except for some that designed more complex architectures.
Hand-crafted design algorithms already show their efficiency in terms of performance but are not modifiable by all users of the fact that they remain skills in case we need to tune them for one specific problem. Same for semi-automatic methods that present complex architecture and lot of hyper-parameter that must be understood before evolving. Standard methods despite their efficiency have not been adaptable all-time in local context problematic \cite{youme_deep_2021}. In recent years, automatic search architectures have shown their efficiency with the architecture cited above in accuracy and performance. Automatic search has many advantages: adaptability to the local context, modifiable according to our goal. However, they need a lot of computer resources around 2500 GPU a Day. This is not available to all scientific around the world. For example, in some sub-Saharan African countries, we don't have this capacity of computer resources despite the fact that some states, in cooperation with foreign organizations, have set up supercomputers. Therefore it's important for us to design an automatic search that keeps both: performance and acceptable cost resources. Cenet presents also simplicity in architecture design and can be easily edited to an other architecture context like natural language processing.

\section{Conclusion}

This paper proposes a genetic algorithm framework for automatically designing an architecture to find the best CNN model in low-cost resources and adaptable to any situation. For this, we have set up a framework with evolution under length constraints. The framework is made possible by two main algorithms: A search length strategy and a search architecture strategy. The first algorithm searches for the optimal space where we should be able to find the best individual by dividing the space into several sub-spaces. The optimal spaces are returned to the next algorithm. This allows us to save search spaces for the genetic algorithm and reduce the computational time. An evolutionary algorithm based on a genetic algorithm which is the most widely used type of algorithm in the field of evolution is used to find the best architecture. To show the effectiveness of this algorithm, we computed it in a benchmark dataset and compared the final result with our competitors. As compared to current peer competitors, our framework presents good performances and lower costs. Cenet's architecture is simple, it is easily scalable to integrate other CNN architectures like object detection architectures. Our next work will be to implement this framework and compute the context of local problems.


%

\begin{acks}

This publication was made possible through the DSTN supported by IRD and AFD. We would like to thank the African Center of Excellence in Mathematical, Informatics, and Tics (CEA-MITIC) and the African Centre of Excellence in Mathematical Science Informatics and their Application (CEA-SMIA) for their support.

\end{acks}



%

\bibliography{MyLibrary}

\end{document}